%% file: main.tex
\documentclass[10pt,twocolumn,letterpaper]{article}

\usepackage{cvpr}              % To produce the CAMERA-READY version
\input{preamble}

\definecolor{cvprblue}{rgb}{0.21,0.49,0.74}
\usepackage[pagebackref,breaklinks,colorlinks,allcolors=cvprblue]{hyperref}

\def\paperID{} % *** Enter the Paper ID here
\def\confName{CVPR}
\def\confYear{2026}

\title{Fose: Fusion of One-Step Diffusion and End-to-End Network for Pansharpening}

\author{
  Kai Liu$^{1}$\thanks{Equal contribution}~,\enspace
  Zeli Lin$^{1}$\footnotemark[1]~,\enspace
  Weibo Wang$^{1}$,\enspace
  Linghe Kong$^{1}$\thanks{Corresponding authors: Linghe Kong,  linghe.kong@sjtu.edu.cn}~,\enspace
  Yulun Zhang$^{1}$\\
  \textsuperscript{1}Shanghai Jiao Tong University.
}

\begin{document}
\maketitle

\setlength{\abovedisplayskip}{2pt}
\setlength{\belowdisplayskip}{2pt}

\begin{abstract}
Pansharpening is a significant image fusion task that fuses low-resolution multispectral images (LRMSI) and high-resolution panchromatic images (PAN) to obtain high-resolution multispectral images (HRMSI).
The development of the diffusion models (DM) and the end-to-end models (E2E model) has greatly improved the frontier of pansharping.
DM takes the multi-step diffusion to obtain an accurate estimation of the residual between LRMSI and HRMSI.
However, the multi-step process takes large computational power and is time-consuming.
As for E2E models, their performance is still limited by the lack of prior and simple structure.
In this paper, we propose a novel four-stage training strategy to obtain a lightweight network Fose, which fuses one-step DM and an E2E model.
We perform one-step distillation on an enhanced SOTA DM for pansharping to compress the inference process from 50 steps to only 1 step.
Then we fuse the E2E model with one-step DM with lightweight ensemble blocks.
Comprehensive experiments are conducted to demonstrate the significant improvement of the proposed Fose on three commonly used benchmarks.
Moreover, we achieve a 7.42 speedup ratio compared to the baseline DM while achieving much better performance.
The code and model will be released soon.
\end{abstract}

\section{Introduction}
Pansharpening is an image fusion technique that integrates a low-resolution multispectral image (LRMSI) with a high-resolution single-band panchromatic (PAN) image to produce a high-resolution multispectral product (HRMSI). 
Owing to physical and cost constraints, spaceborne multispectral and especially hyperspectral sensors typically offer limited spatial resolution that falls short of many downstream application requirements. 
In contrast, mounting a PAN sensor on the same platform is comparatively inexpensive. 
This sensor configuration makes pansharpening a de facto capability for modern satellites and a crucial enabler for numerous tasks, including motion detection, change detection~\cite{wu2017post}, and semantic segmentation~\cite{yuan2021review}.

\begin{figure}[t]
    \centering
    \includegraphics[width=\linewidth]{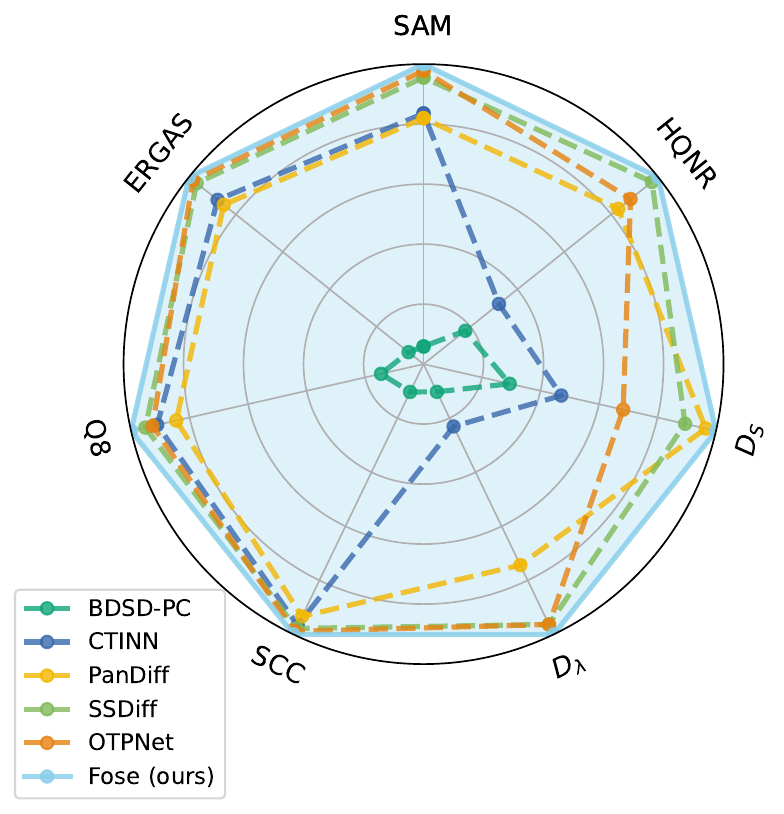}
    \caption{Performance comparison with SOTA methods on WV3. Fose achieves robust performance across all metrics.}
    \label{fig:radar}
\end{figure}

Pansharpening dates back to the 1970s~\cite{kwarteng1989extracting}. 
Classical approaches are commonly grouped into three families: component substitution (CS), multi-resolution analysis (MRA), and variational optimization (VO). 
CS methods project the low-resolution LRMSI into a transform domain and replace its spatial component(s) with those extracted from the PAN image, yielding high spatial fidelity but often introducing spectral distortions while being computationally efficient~\cite{kwarteng1989extracting,meng2016pansharpening}. MRA methods perform multiscale decompositions to extract spatial details from the PAN image and inject them into the LRMSI, generally preserving spectral information at the possible expense of spatial sharpness~\cite{otazu2005introduction,vivone2017regression}. VO formulations provide stronger mathematical guarantees than CS and MRA but typically incur higher computational cost and require careful tuning of multiple hyperparameters~\cite{wu2023lrtcfpan,wu2023framelet}.

With the rise of deep learning, data-driven pansharpening has advanced rapidly. 
PNN was the first model to introduce deep learning to pansharpening~\cite{masi2016pansharpening}.
It upsamples the LRMSI to the PAN resolution, concatenates it with the PAN image along the channel dimension.
Then it learns a supervised mapping to the ground-truth high-resolution MSI through a three-layer CNN.
Subsequent end-to-end architectures take LRMSI and PAN as inputs and produce HRMSI in a single forward pass, offering low parameter counts and modest computational demands, albeit with limited peak performance.
Scaling such networks by simply increasing width or depth often yields diminishing returns and, in practice, can degrade generalization due to overfitting. 
Diffusion-based models, by contrast, predict the residual between LRMSI and HRMSI, typically conditioning on the PAN image.
Through a multi-step denoising process, they convert noise into the residual and have shown clear advantages in spatial detail and visual fidelity. 
However, their iterative sampling is computationally heavy, often one to two orders of magnitude slower than end-to-end models.

To reconcile these trade-offs, we propose Fose, which distills a multi-step diffusion model into a single-step generator and then fuses it with an end-to-end network via a lightweight convolutional adaptor.
Concretely, we first strengthen a state-of-the-art diffusion-based pansharpening baseline by integrating adaptive convolution, achieving a strong reference model without a substantial parameter increase.
We then perform single-step distillation using a VSD loss, accelerating inference by roughly 50× with negligible loss in accuracy.
Then, we train a typical end-to-end model to compensate for the degradation from the 50x compression.
Finally, we fuse the outputs of the distilled one-step diffusion model and the end-to-end model using a shallow convolutional fusion head. 
During training of this fusion stage, the backbone models are frozen and only the fusion parameters are updated, enabling rapid convergence and performance that consistently surpasses either component alone.
Our contributions are threefold:

\begin{itemize}
    \item We propose Fose, the first one-step diffusion model for pansharping, fused with an end-to-end network. Fose obtains a 7.42 $\times$ speedup ratio compared to its multi-step baseline model while achieving better performance. 

    \item We propose a four-stage training strategy to gradually improve the performance and the speedup ratio. Generally, the strategy focuses on obtaining a strong baseline model for both the DM and E2E model, one-step distillation, and fusing them with light ensemble connector layers.

    \item We conduct comprehensive experiments on three commonly used pansharpening datasets to validate the excellent performance compared to previous SOTA models. Also, the ablation study shows the effectiveness and robustness of the proposed Fose. 
    
\end{itemize}

\section{Related Works}
\subsection{DL-based Methods}
As a simple yet effective strategy, the representative single-scale coupling model exemplified by PNN employs a compact three-layer CNN that achieved state-of-the-art performance at the time. 
Subsequently, methods such as FusionNet~\cite{Deng2020FusionNet} and DCFNet~\cite{Wu2021DCFNet} adopted similar coupled-input designs.
However, due to the limited feature representation capacity of these architectures, their performance in terms of spectral fidelity and generalization remains insufficient. 
In multi-source image fusion, inputs acquired from heterogeneous sensors exhibit distinct characteristics, and directly coupling the two sources often leads to suboptimal feature extraction. 
In contrast, deep learning methods with separate spatial and spectral branches can more effectively discriminate and hierarchically learn information from PAN images and LRMSI, thereby better exploiting multi-scale information.
As a successful practice, OTPNet incorporates the concept of ODEs to construct an efficient architecture that eliminates the need for complex tuning~\cite{yu2025otpnet}.
Despite being efficient and effective, these methods quickly reach the performance bound with increasing model size.
Therefore, pansharpening needs new structures and fusion methods to achieve better performance.

\subsection{Diffusion-based Models}
Denoising Diffusion Probabilistic Models (DDPMs) have been widely adopted in tasks such as text-to-image generation and image editing, and have recently demonstrated strong performance in image processing~\cite{ho2020denoising}. 
Building on DDPMs, the Denoising Diffusion Implicit Model (DDIM) proposed by Song et al~\cite{song2020denoising} introduces a non-Markovian sampling scheme that substantially accelerates inference. 
Further along this line, Improved DDPM (IDDPM) attains competitive log-likelihoods while preserving the high sample quality of DDPM through several simple architectural and training refinements~\cite{nichol2021improved}. 
Recently, DDPMs have also attracted increasing interest in pansharpening.
Instead of directly fusing PAN and LRMSI as in conventional methods, DDPM-based approaches treat them as conditioning signals for generative fusion.
However, diffusion models typically incur computational costs that are an order of magnitude higher than those of end-to-end models.
Recently, research on one-step diffusion models has begun to emerge~\cite{wu2024one}. 
These methods take the multi-step diffusion models as the teacher to distill a one-step diffusion model with negligible performance degradation~\cite{wang2025osdface}.
These distillation methods usually take an additional lightweight module to compensate for the degradation from the extreme compression.
To the best of our knowledge, we are the first to apply a one-step diffusion model to pansharpening, achieving substantial acceleration while further improving performance.

\begin{figure*}[t]
    \centering
    \includegraphics[width=0.90\linewidth]{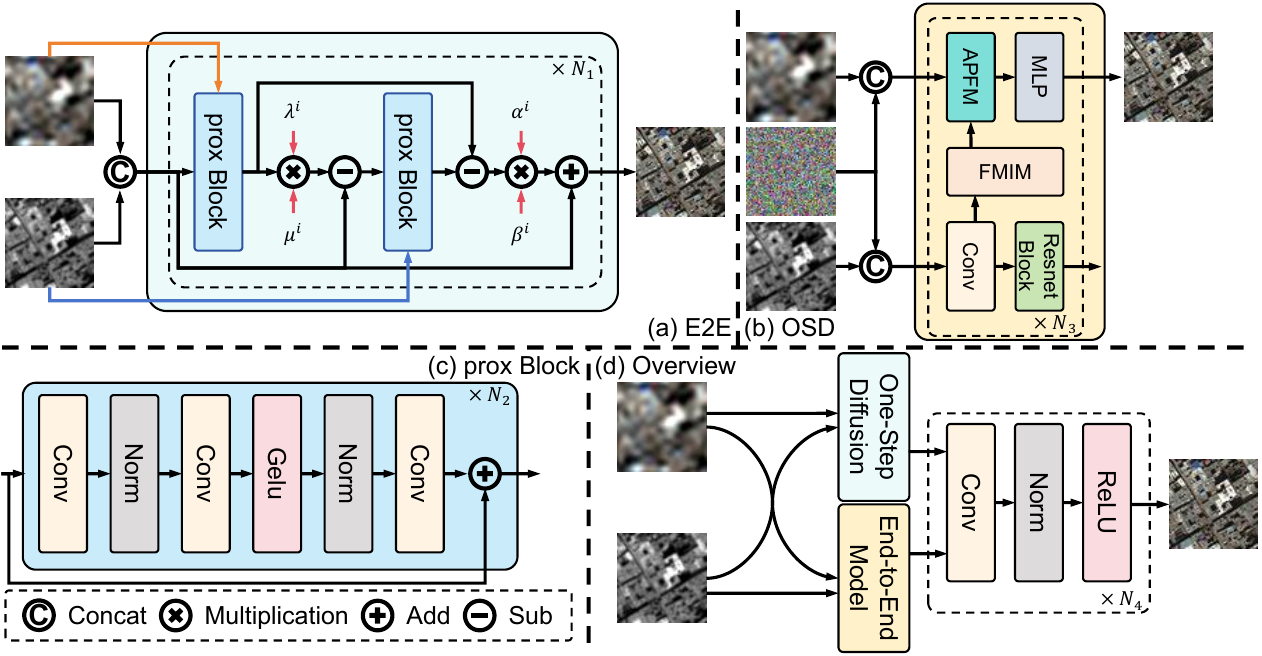}
    \caption{Architecture of the proposed Fose. Fose consists of three components, including the OSD, E2E model, and an ensemble connector. The E2E model adopts an ODE proximal network structure, which obtains excellent performance with a relatively small size. The OSD model consists of two branches to process MSI and PAN separately and fuse them with APFM, an attention-like structure. The ensemble connector network leverages lightweight convolutional blocks to fuse the output of two models to obtain the final output.}
    \label{fig:architecture}
    \vspace{-4mm}
\end{figure*}

\section{Method}
\subsection{Preliminary}
Denoising diffusion probabilistic models (DDPMs)~\cite{ho2020denoising} are latent-variable generative models that progressively synthesize realistic target images from a standard normal prior by iterative denoising. A diffusion model comprises two phases: a forward (noising) process and a reverse (denoising) process.

In the forward process, the clean data $\mathbf{x}_0$ are gradually perturbed through a $T$-step Markov chain until they approach a standard normal distribution, i.e., $\mathbf{x}_T \sim \mathcal{N}(\mathbf{0}, \mathbf{I}_d)$, where $d$ is the dimensionality. One step of the chain is
\begin{align}
q(\mathbf{x}_{t}\mid \mathbf{x}_{t-1})
= \mathcal{N}\bigl(\mathbf{x}_t;\sqrt{1-\beta_t},\mathbf{x}_{t-1}, \beta_t \mathbf{I}\bigr),
\label{eq: forstep}
\end{align}
where $\beta_t\in(0,1)$ is a predefined variance schedule for $t\in\{1,\dots,T\}$ and $\mathcal{N}(\cdot)$ denotes a Gaussian distribution with the indicated mean and covariance. 
Using the reparameterization trick, $\mathbf{x}_t$ can be written in closed form as a noisy version of $\mathbf{x}_0$:
\begin{align}
\mathbf{x}_{t}
= \sqrt{\bar{\alpha}_t}, \mathbf{x}_{0} + \sqrt{1-\bar{\alpha}_t},\boldsymbol{\epsilon},
\qquad
\boldsymbol{\epsilon}\sim \mathcal{N}(\mathbf{0},\mathbf{I}),
\label{eq: forstep1}
\end{align}
with $\alpha_t = 1-\beta_t$ and $\bar{\alpha}_t=\prod_{i=1}^{t}\alpha_i$.

The reverse process learns to invert the forward corruption and recover $\mathbf{x}_0$ from a noisy sample $\mathbf{x}_t$. To this end, a neural network parameterized by $\theta$ models the transition
\begin{align}
p_\theta(\mathbf{x}_{t-1}\mid \mathbf{x}_{t})
= \mathcal{N}\bigl(\mathbf{x}_{t-1}; \boldsymbol{\mu}_\theta(\mathbf{x}_t,t), \boldsymbol{\Sigma}_\theta(\mathbf{x}_t,t)\bigr),
\label{eq: revprocess}
\end{align}
where $\boldsymbol{\mu}_\theta$ and $\boldsymbol{\Sigma}_\theta$ denote the mean and (typically diagonal) covariance, respectively, and $\theta$ are the model parameters.

Following \eqref{eq: revprocess}, the mean and variance used at step $t$ can be computed as
\begin{align}
\boldsymbol{\mu}_\theta(\mathbf{x}_t,t)
= \frac{1}{\sqrt{\alpha_t}}
\left(\mathbf{x}_t - \frac{\beta_t}{\sqrt{1-\bar{\alpha}_t}},
\boldsymbol{\epsilon}_\theta(\mathbf{x}_t,t)\right),
\label{eq: mu}
\end{align}
\begin{align}
\boldsymbol{\Sigma}_\theta(\mathbf{x}_t,t)
= \frac{1-\bar{\alpha}_{t-1}}{1-\bar{\alpha}_{t}}\beta_t.
\label{eq: sigma}
\end{align}
Starting from $\mathbf{x}_T \sim \mathcal{N}(\mathbf{0},\mathbf{I}_d)$, iterating the reverse transitions for $t=\{T,\dots,1\}$ yields a sample $\mathbf{x}_0$ from the learned data distribution.

\subsection{Architecture and Training Strategy}
The overall architecture of the proposed Fose can be divided into three parts, including the one-step diffusion model, end-to-end model, and the ensemble connector, as shown in Fig.~\ref{fig:architecture}.
We adopt a four-stage training strategy to optimize the network, including (1) improving the multi-step diffusion model, (2) one-step distillation, (3) end-to-end training, and (4) model fusion.
We then thoroughly describe the strategy and the corresponding architecture.

\noindent\textbf{Stage 1.}
The core objective of Stage 1 is to establish a baseline model by using an SOTA multi-step diffusion model for pansharping.
With additional modules aimed at further enhancing the model's performance, a strong foundation for the subsequent single-step distillation is thereby provided.
Specifically, we adopt SSDiff~\cite{zhong2024ssdiff} as our baseline model. 
It incorporates spectral and spatial branches to separately process LRMSI and panchromatic images. 
Within each sub-module, the high-frequency components of the panchromatic image are extracted with a Fourier mask within the FMIM~\cite{si2024freeu}.
The feature from the MSI branch and the panchromatic branch is fused with an attention-like structure~\cite{zhong2024ssdiff}.
Ultimately, the input noise is optimized into the residual between the HRMSI and LRMSI through a multi-step diffusion process, which is then added to the LRMSI.

However, unlike general computer tasks, the scales of key targets vary significantly, ranging from a few pixels to nearly the entire image.
Consequently, the use of standard convolution kernels somewhat limits the performance of the diffusion model. 
To address this, we apply the ARConv~\cite{wang2025adaptive}, utilizing adaptive-scale convolution kernels and perceptual range to enhance the model's ability to perceive objects at different scales.
The convolution process is
\begin{equation}
    \mathbf{y}\left(\mathbf{p}_{0}\right)=\sum_{\mathbf{r}_{n} \in \mathbf{R}} \mathbf{w}\left(\mathbf{r}_{n}\right) \cdot \mathbf{t}\left(\mathbf{p}_{0}+\mathbf{r}_{n}\right),
\end{equation}
where $\mathbf{y}\left(\mathbf{p}_{0}\right)$ refers to the pixel value at position $\mathbf{p}_{0}$ in the output feature map $\mathbf{y}$, $\mathbf{w}$ denotes the weight of the convolution layer, $\mathbf{r}_{n}$ enumerates the elements in $\mathbf{R}$, $\mathbf{t}\left(\mathbf{p}_{0}+\mathbf{r}_{n}\right)$ calculates the pixel value with interpolation at position $\mathbf{p}_{0}+\mathbf{r}_{n}$.
Both the choice of $\mathbf{w}$ and $\mathbf{R}$ are dynamically determined by a light residual block during the first $T$ training epochs.
After $T$ epochs, the choice of $\mathbf{w}$ will be fixed while its weight is still being optimized.

We replace all the convolution layers within Resblocks of SSDiff~\cite{zhong2024ssdiff} with ARConv~\cite{wang2025adaptive}.
Despite the additional parameters being introduced, the performance improvement is worthwhile.
The optimization objective for this stage follows the standard diffusion model optimization goal~\cite{cao2024diffusion}:
\begin{equation}
    \mathcal{L}_{S1}=\mathbb{E}\left[\left\|x_{0}-x_{\theta}\left(\mathbf{x}_{t}, \mathbf{c}, t\right)\right\|_{1}\right],
\end{equation}
where $x_{\theta}$ denotes the prediction of the diffusion model, $\mathbf{c}$ is the conditions for injecting the model, \ie, the LRMSI and panchromatic image, and $\left\|\cdot\right\|$ denotes L1 norm.

\noindent\textbf{Stage 2.} 
The core objective of Stage 2 is to distill the diffusion model from a multi-step to a single-step model.
With the single-step distillation, the inference speed can be significantly accelerated while ensuring that performance does not degrade substantially. 
We use VSD loss $\mathcal{L}_{\text{VSD}}$ and data loss $\mathcal{L}_{\text{data}}$ to distill the diffusion model~\cite{wang2023prolificdreamer}:
\begin{equation}
    \mathcal{L}_{\text{VSD}} = \mathbb{E}_{t} \left[
    \frac{1}{\mathbb{E}\!\left[ \lVert x_{0}^{T} - x_{0}^{S} \rVert \right]}\cdot \lVert x_{0}^{T} - x_{0}^{S} \rVert^{2}
\right],
\end{equation}
while the data loss is expressed as:
\begin{equation}
    \mathcal{L}_{\text {data }}=\left\| G_{\theta}\left(\boldsymbol{x}_{L}\right)-\boldsymbol{x}_{H} \right\|_1.
\end{equation}
Here, we exclusively use pixel loss instead of perceptual losses such as LPIPS~\cite{ghildyal2022shift}, CLIP-IQA~\cite{wang2023exploring}, and others, due to the distinct data distribution patterns in remote sensing domains.
While these perceptual losses perform well on natural images, they are not suitable for remote sensing imagery. 
We provide the pseudo code in Alg.~\ref{alg:train-stage2} to demonstrate the one-step distillation process more precisely.

\noindent\textbf{Stage 3.} 
Stage 3 focuses on training an end-to-end network. 
For this part, we follow OTPNet~\cite{yu2025otpnet}, a DL model specifically designed for RS image fusion tasks.
It is inspired by numerical solution methods of ordinary differential equations (ODEs). 
The network incorporates the concept of ODEs to construct an efficient architecture that eliminates the need for complex tuning, aiming to enhance both the performance and stability of remote sensing image fusion.

The core idea of OTPNet is to leverage numerical methods of ODEs, such as Euler’s method and Runge-Kutta methods, to design the network structure.
This approach simulates the dynamic process of ODEs, optimizing the network's local truncation error, thereby achieving more efficient feature extraction and fusion.
The training loss function here employs the simple and effective L1 loss $\mathcal{L}_{S3}=\mathbb{E}\left[\left\|x_{0}-x_{\theta}\left(\mathbf{x}_{t}, \mathbf{c}, t\right)\right\|_{1}\right]$.

\noindent\textbf{Stage 4.} 
Stage 4 integrates a lightweight convolution layer to fuse the outputs of the OSD model and the E2E model.
The ensemble connector first concatenates the output of OSD $\mathbf{y}_{OSD}$ and end-to-end model $\mathbf{y}_{e}$ channel-wise:
\begin{equation}
    \mathbf{x}_{c} = \text{Concat}(\mathbf{y}_{OSD}, \mathbf{y}_{e}).
\end{equation}
Then, we leverage lightweight convolution layers to further enhance the performance.
For each ensemble block, the process can be written as:
\begin{equation}
    \mathbf{x}_{c}^{\prime} = \text{ReLU}(\text{BN}(\text{Conv}(\mathbf{x}_{c}))).
\end{equation}
We freeze the weight of the one-step diffusion model and the end-to-end model.
By optimizing the output of connector $\mathbf{y}_c$ with $\mathcal{L}_{S4}=\mathbb{E}\left[\left\|\mathbf{y}_c-\mathbf{y}\right\|_{1}\right]$, the final performance can be greatly improved.

\begin{algorithm}
\SetKwInOut{Input}{Input}\SetKwInOut{Output}{Output}
\caption{One-step distillation.}
\label{alg:train-stage2}
\Input{Ground truth image $X_0$, pretrained multi-step diffusion model $\epsilon_\theta$ with parameters $\theta$, condition $\text{cond}$, timestep $t$,  training itration $N$}
\Output{One-step distilled diffusion model $\epsilon^*_\theta$ }
 Initialize $\epsilon^*_\theta \leftarrow \epsilon_\theta$ with trainable LoRA\;
$\text{cond} \leftarrow \{\text{PAN}, \text{LRMSI}, X_t\}$\;
\For{$i \leftarrow 1$ TO $N$}{
    $t \sim \mathcal{U}(20, 980)$\;
    $\epsilon \sim \mathcal{N}(0, I)$\;
    $X_t \leftarrow \sqrt{\bar{\alpha}_t}(X_0 - \text{LRMSI}) + \sqrt{1 - \bar{\alpha}_t}\,\epsilon$\;
    $\hat{x}_0 ^{T} \leftarrow \text{stopgrad }\epsilon_\theta(X_t, \text{cond}) + \text{LRMSI}$\;
    $\hat{x}_0 ^{S} \leftarrow \epsilon^*_\theta(X_t, \text{cond}) + \text{LRMSI}$\;
    $\omega \leftarrow 1/\mathrm{mean}(\|\hat{x}_0 ^{T} - \hat{x}_0 ^{S}\|)$\;
    $\mathcal{L}_{vsd} \leftarrow \omega \cdot \left\lVert \hat{x}^{T}_{0} - \hat{x}^{S}_{0} \right\rVert^2$\;
    $\mathcal{L}_{data} \leftarrow \mathcal{L}_{1}(\hat{x}_0 ^{S},X_0)$\;
    Update $\theta^*$ with $\mathcal{L}_{vsd}+\mathcal{L}_{data}$\;
}
\end{algorithm}

\begin{table*}[t]
\centering
\caption{Experimental results on WV3. We compare our proposed Fose with both E2E models and diffusion models. Both reduced resolution and full resolution metrics are adopted to thoroughly evaluate the performance of Fose. In conclusion, Fose achieves obvious SOTA performance over all metrics compared to previous SOTA pansharpening methods.}
\vspace{-2mm}
\label{tab:exp-main-comp}
\resizebox{\linewidth}{!}{
\begin{tabular}{l|cccc:ccc}
\toprule[0.15em]
\rowcolor{table_gray}   & \multicolumn{4}{c}{RR}                                                                     & \multicolumn{3}{c}{FR}                                         \\
\rowcolor{table_gray}   \multirow{-2}{*}{Method}    & SAM $\downarrow$     & ERGAS $\downarrow$ & Q8/Q4 $\uparrow$          & SCC $\uparrow$        & $D_\lambda$ $\downarrow$ & $D_S$ $\downarrow$  & HQNR $\uparrow$   \\
\midrule[0.05em]
\midrule[0.05em]
BDSD-PC    & 5.468±1.718          & 4.655±1.467        & 0.812±0.106            & 0.905±0.042           & 0.062±0.024       & 0.073±0.036           & 0.870±0.053        \\
MTF-GLP-FS & 5.323±1.655          & 4.645±1.444        & 0.818±0.101            & 0.898±0.047           & 0.021±0.008       & 0.063±0.028           & 0.918±0.035        \\
BT-H       & 4.899±1.303          & 4.515±1.331        & 0.818±0.102            & 0.924±0.024           & 0.057±0.023       & 0.081±0.037           & 0.867±0.054        \\
PNN        & 3.680±0.762          & 2.682±0.647        & 0.893±0.092            & 0.976±0.007           & 0.021±0.008       & 0.043±0.015           & 0.937±0.021        \\
DiCNN      & 3.593±0.762          & 2.673±0.663        & 0.900±0.087            & 0.976±0.007           & 0.036±0.011       & 0.046±0.018           & 0.919±0.026        \\
MSDCNN     & 3.777±0.803          & 2.761±0.688        & 0.890±0.090            & 0.974±0.008           & 0.023±0.009       & 0.047±0.020           & 0.932±0.027        \\
FusionNet  & 3.325±0.698          & 2.467±0.645        & 0.904±0.090            & 0.981±0.007           & 0.024±0.009       & 0.036±0.014           & 0.941±0.020        \\
CTINN      & 3.252±0.644          & 2.394±0.519        & 0.906±0.091            & 0.983±0.005           & 0.055±0.029       & 0.068±0.031           & 0.881±0.049        \\
LAGConv    & 3.104±0.558          & 2.300±0.613        & 0.910±0.091            & 0.984±0.007           & 0.037±0.015       & 0.042±0.015           & 0.923±0.025        \\
MMNet      & 3.084±0.640          & 2.343±0.626        & 0.915±0.089            & 0.983±0.006           & 0.054±0.023       & 0.034±0.011           & 0.914±0.028        \\
DCFNet     & 3.026±0.740          & 2.159±0.456        & 0.905±0.088            & 0.986±0.004           & 0.078±0.081       & 0.051±0.034           & 0.877±0.101        \\
PanDiff    & 3.297±0.601          & 2.467±0.584        & 0.898±0.081            & 0.980±0.006           & 0.027±0.012       & 0.054±0.026           & 0.920±0.036        \\
ADWM       & 2.900±0.507          & 2.136±0.474        & 0.917±0.080            & 0.984±0.005           & 0.020±0.008       & 0.067±0.037           & 0.915±0.042        \\
SSDiff     & 2.910±0.505          & 2.146±0.459        & 0.911±0.086            & 0.984±0.005           & 0.015±0.008       & 0.056±0.031           & 0.931±0.038        \\
OTPNet     & 2.846±0.495          & 2.102±0.469        & 0.908±0.094            & 0.985±0.004           & 0.015±0.008       & 0.062±0.032           & 0.924±0.038        \\
\rowcolor{table_green} Fose (ours)       & 2.783±0.504        & 2.055±0.445            & 0.917±0.082           & 0.986±0.004       & 0.013±0.008           & 0.053±0.033           & 0.933±0.037        \\
\bottomrule[0.15em]
\end{tabular}
} % resize box
\vspace{-5mm}
\end{table*}

\section{Experiments}
\subsection{Implementation Detail}

Our Fose framework is implemented in Python~3.10 and PyTorch~1.7.0~\cite{paszke2019pytorch}.
We adopt the AdamW~\cite{loshchilov2017decoupled} optimizer. 
All experiments are conducted on a Linux operating system equipped with a single NVIDIA RTX~3090 GPU. 

For the diffusion denoising model, the initial number of model channels is set to $128$. 
In the pansharpening task, the diffusion time step for training is fixed at $1000$ at the first stage, while the time step used for sampling is set to $50$. 
The total number of training iterations for the WV3, GF2, and QB datasets is configured as $150\text{k}$, $300\text{k}$, and $350\text{k}$, respectively. 
The overall training process is organized into four stages, whose batch sizes are set to $64$, $16$, $24$, and $16$ from the first to the fourth stage.
The corresponding initial learning rates for these four stages are $1 \times 10^{-4}$, $2 \times 10^{-4}$, $5 \times 10^{-5}$, and $1 \times 10^{-4}$, respectively.
More detailed settings can be found in the supplementary material.

\subsection{Datasets}
To comprehensively evaluate the effectiveness of our method, we conduct experiments on multiple widely used remote sensing datasets.
Specifically, we consider 8-band multispectral imagery provided by the WorldView-3 (WV3) sensor, as well as 4-band data acquired by the QuickBird (QB) and GaoFen-2 (GF2) sensors.
Since HRMSI ground truth is unavailable for these satellites, we follow the standard Wald's protocol~\cite{wald1997fusion} to generate training and testing pairs at reduced resolution.
All datasets used in this study are publicly accessible through an open repository.

\subsection{Evaluation}
To evaluate the performance of our Fose, we compare it with various state-of-the-art methods of pansharping.
To be specific, we choose three traditional methods, including BDSD-PC~\cite{Vivone2019BDSDPC}, MTF-GLP-FS~\cite{Vivone2018MTF}, and BTH~\cite{Aiazzi2006BTH}.
In addition, the remaining twelve approaches are representative deep learning–based pansharpening methods.
These approaches can be broadly categorized into two groups.

The first is CNN-based models, including PNN~\cite{Masi2016PNN}, DiCNN~\cite{He2019DiCNN}, MSDCNN~\cite{Wei2017MSDCNN}, FusionNet~\cite{Deng2020FusionNet}, CTINN~\cite{Zhou2022CTINN}, LAGConv~\cite{Jin2022LAGConv}, MMNet~\cite{Zhou2023MMNet}, OTPNet~\cite{yu2025otpnet}, and DCFNet~\cite{Wu2021DCFNet}.
The second part is diffusion-based generative models, including PanDiff~\cite{Meng2023PanDiff}, ADWM~\cite{Huang2025ADWM}, SSDiff~\cite{zhong2024ssdiff}.

\subsection{Metrics}
Given the variation in spatial resolution across datasets, we adopt different quantitative indicators for reduced and full-resolution evaluations.
For reduced resolution (RR), we adopt 4 metrics, including SAM~\cite{Boardman1993AVIRIS}, ERGAS~\cite{Wald2002FusionDefinition}, Q8~\cite{Garzelli2009Q8}, and SCC~\cite{Zhou1998Wavelet}.
For full resolution (FR), we adopt 3 metrics, including $D_s$, $D_\lambda$, and HQNR~\cite{Arienzo2022HQNR}.

\subsection{Experimental Results}

\begin{figure*}[t]
    \centering
    \includegraphics[width=0.95\linewidth]{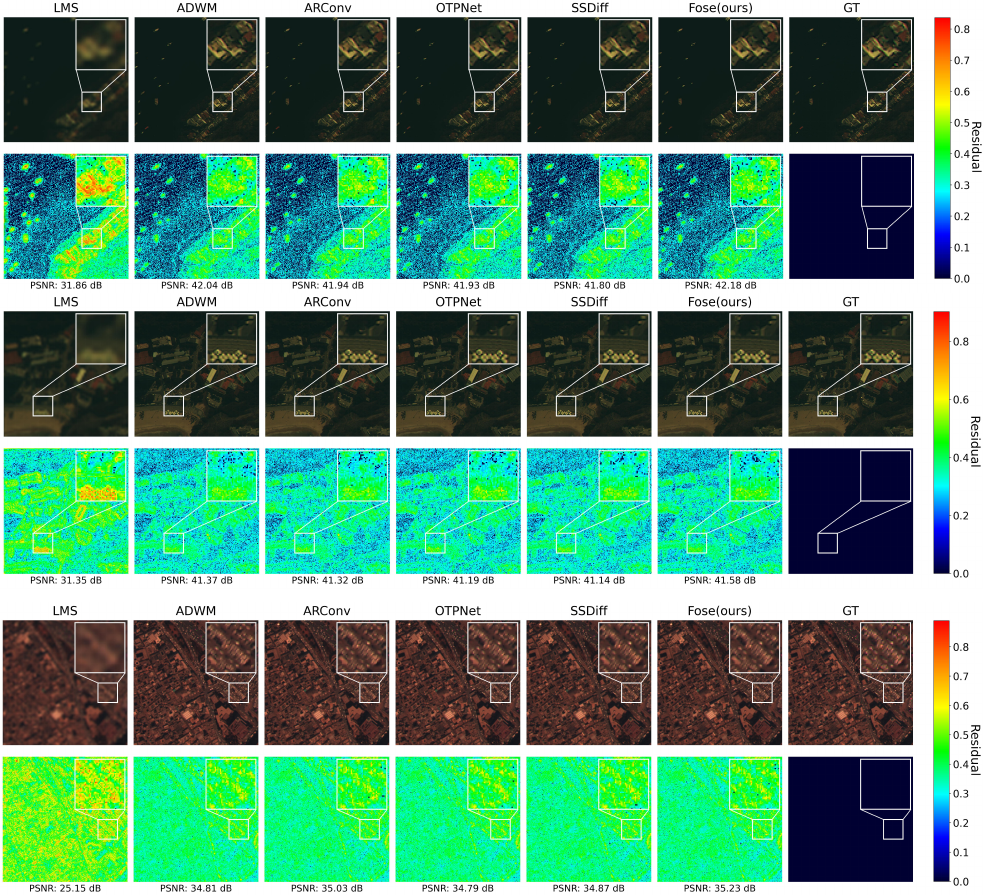}
    \caption{Visual comparison with SOTA pansharpening methods. We provide both the MSI and the residual between the fused image and the GT for better visualization. The average residual of Fose is visually smaller than other methods, validating the effectiveness of Fose.}
    \label{fig:visual-comparison}
\end{figure*}

\noindent\textbf{Quantitative Comparison.}
The quantitative evaluation demonstrates the superior performance of the proposed method across comprehensive assessment metrics for pansharpening quality.
As illustrated in Tab.~\ref{tab:exp-main-comp}, a clear performance hierarchy emerges, with deep learning-based approaches substantially outperforming traditional component substitution and multiresolution analysis methods. 
The conventional techniques, including BDSD-PC, MTF-GLP-FS, and BT-H, exhibit the highest relative dimensionless global errors (RR \> 4.5) and feature reconstruction errors (FR \> 4.5), coupled with elevated spectral distortion measures. 
This performance gap underscores the limitation of hand-crafted filters in simultaneously preserving spatial detail and spectral fidelity.
Among the convolutional neural network architectures, the proposed method establishes a new state-of-the-art, achieving the lowest mean errors in both RR (2.783±0.504) and FR (2.055±0.445) metrics, surpassing the previous best-performing OTPNet by 2.2\% and 2.2\%, respectively. 
The statistical significance is evident, as the standard deviations indicate consistent performance across test samples. 
The Spectral Angle Mapper (SAM) values reveal that the proposed method maintains excellent spectral fidelity (0.917±0.082), comparable to ADWM (0.917±0.080) and MMNet (0.915±0.089), while substantially improving spatial reconstruction. 
The ERGAS metric further corroborates this advancement, where the proposed method attains 0.986±0.004, ranking among the top and demonstrating minimal spectral-spatial distortion.

\begin{table*}[t]
\centering
\caption{Structure and strategy ablation on WV3. The ablation proves the effectiveness of the proposed four-stage training strategy.}
\vspace{-2mm}
\label{tab:module-ablation}
\resizebox{\linewidth}{!}{
\begin{tabular}{l|l|ccrr|ccc|cc}
\toprule[0.15em]
Row  & \multirow{2}{*}{Method} & \multirow{2}{*}{Step} & Ensenble & FLOPs  & Params  & \multicolumn{2}{c}{RR}                                & \multicolumn{2}{c}{FR}               \\
Ann. &                         &                       & Layer    & (G)    & (M)     & SAM $\downarrow$ & ERGAS $\downarrow$ & Q8 $\uparrow$  & $D_S$$\downarrow$ & HQNR $\uparrow$ \\
\midrule[0.05em]
R1   & DM                      & 50  & \xmark & 2205.00 & 1.43 & 2.910±0.505 & 2.146±0.459 & 0.911±0.086 & 0.056±0.031 & 0.931±0.038 \\
R2   & DM+ARConv               & 50  & \xmark & 6495.00 & 3.63 & 2.885±0.515 & 2.123±0.440 & 0.910±0.086 & 0.051±0.030 & 0.930±0.037 \\
\midrule[0.05em]
R3   & E2E Model               & 1   & \xmark &  154.73 & 2.36 & 2.846±0.495 & 2.102±0.469 & 0.908±0.094 & 0.062±0.032 & 0.924±0.038 \\
R4   & DM                      & 1   & \xmark &   45.17 & 1.44 & 2.933±0.512 & 2.174±0.465 & 0.911±0.085 & 0.057±0.032 & 0.930±0.038 \\
R5   & DM+ARConv               & 1   & \xmark &  130.03 & 3.64 & 2.903±0.520 & 2.148±0.449 & 0.909±0.086 & 0.053±0.031 & 0.929±0.039 \\
R6   & E2E Model               & 1   & \cmark &  167.25 & 2.56 & 2.823±0.509 & 2.089±0.465 & 0.919±0.080 & 0.064±0.032 & 0.923±0.038 \\
R7   & DM+ARConv               & 1   & \cmark &  142.54 & 3.83 & 2.863±0.518 & 2.126±0.447 & 0.912±0.085 & 0.053±0.031 & 0.930±0.038 \\
\midrule[0.05em]
R8   & Fose (ours)             & 1   & \cmark &  297.60 & 6.20 & 2.783±0.504 & 2.055±0.445 & 0.917±0.082 & 0.053±0.033 & 0.933±0.037 \\
\bottomrule[0.15em]
\end{tabular}
} % resize box
\vspace{-2mm}
\end{table*}

Comparative analysis with recent diffusion-based approaches reveals nuanced trade-offs. 
While PanDiff achieves competitive SAM and ERGAS scores, its RR (3.297±0.601) and FR (2.467±0.584) remain significantly higher than the proposed method, indicating inferior spatial structure preservation. 
The ADWM method, representing the current state-of-the-art for WorldView-3 imagery, exhibits strong spectral performance but exhibits higher spatial distortion (D = 0.067±0.037) compared to the proposed approach (D = 0.053±0.033). 
This suggests that the integration of attention mechanisms in our architecture more effectively balances the spatial-spectral trade-off.
Similarly, SSDiff and OTPNet, though performing admirably on specific metrics, demonstrate marginal degradations in either SCC or $D_s$ values.

The performance improvement is particularly notable when considering the full quality-no-reference (HQNR) index, where the proposed method achieves 0.933±0.037, exceeding most competitors except PNN (0.937±0.021) and FusionNet (0.941±0.020). 
However, these marginally higher HQNR values for alternative methods do not compensate for their substantially inferior RR and FR performance.
The ablative implication is that optimizing solely for perceptual quality may compromise rigorous radiometric accuracy, a limitation that our multi-objective loss function successfully mitigates.
Furthermore, the reduced variance across most metrics indicates enhanced robustness to diverse landscape categories and atmospheric conditions present in the test dataset.

In summary, the proposed methodology establishes a comprehensive performance advantage, simultaneously minimizing radiometric errors, spectral distortion, and spatial degradation while maintaining competitive no-reference quality scores. 
The consistent improvements across both absolute error metrics and relative quality indices validate the architectural innovations, particularly the adaptive feature fusion module and hierarchical supervision strategy, as effective mechanisms for pansharpening applications.
The results on GF2 and QB can be found in the supplementary material, which further support the robustness of Fose.

\begin{table*}[t]
\centering
\caption{Depth of end-to-end model on WV3. We conduct the depth ablation to demonstrate the limitation of the E2E model. To be specific, by naively increasing the model size, the performance quickly reaches saturation and even begins to degrade.}
\label{tab:ablation-depth}
\resizebox{\linewidth}{!}{
\begin{tabular}{c|rr|cccc|ccc}
\toprule[0.15em]
Number of Blocks     & FLOPs & Params & \multicolumn{4}{c}{RR}                                & \multicolumn{3}{c}{FR}               \\
of E2E Model & (G)   & (M)    & SAM $\downarrow$ & ERGAS $\downarrow$ & Q8 $\uparrow$ & SCC $\uparrow$ & $D_\lambda$$\downarrow$  & $D_S$$\downarrow$ & HQNR $\uparrow$ \\
\midrule[0.05em]
\midrule[0.05em]
3                  & 7.33  & 1.79 & 2.969±0.548 & 2.201±0.462 & 0.914±0.081 & 0.983±0.005 & 0.016±0.013 & 0.080±0.040 & 0.906±0.050 \\
4                  & 9.67  & 2.36 & 2.846±0.495 & 2.102±0.469 & 0.908±0.094 & 0.985±0.004 & 0.015±0.008 & 0.062±0.032 & 0.924±0.038 \\
5                  & 12.00 & 2.94 & 2.901±0.518 & 2.147±0.469 & 0.915±0.081 & 0.984±0.005 & 0.018±0.016 & 0.072±0.038 & 0.912±0.049 \\
6                  & 14.33 & 3.51 & 2.919±0.523 & 2.170±0.469 & 0.914±0.081 & 0.984±0.005 & 0.020±0.022 & 0.074±0.036 & 0.908±0.049 \\
\bottomrule[0.15em]
\end{tabular}
} % resize box
\vspace{-2mm}
\end{table*}

\begin{table*}[h!]
\centering
\caption{Noise ablation on WV3. The choice of initial noise significantly influences the performance of OSD.}
\vspace{-2mm}
\label{tab:ablation-noise}
\begin{tabular}{c|ccccccc}
\toprule[0.15em]
Noise  & SAM $\downarrow$ & ERGAS $\downarrow$ & Q8 $\uparrow$ & SCC $\uparrow$ & $D_\lambda$$\downarrow$  & $D_S$$\downarrow$ & HQNR $\uparrow$ \\
\midrule[0.05em]
\midrule[0.05em]
Zero   & 2.783±0.504        & 2.055±0.445      & 0.917±0.082          & 0.986±0.004         & 0.013±0.008     & 0.053±0.033         & 0.933±0.037      \\
Random & 3.094±0.428        & 2.130±0.455      & 0.866±0.118          & 0.986±0.004         & 0.015±0.010     & 0.064±0.032         & 0.921±0.039     \\
\bottomrule[0.15em]
\end{tabular}
\vspace{-2mm}
\end{table*}

\noindent\textbf{Qualitative Comparison.}
The visual comparison substantiates the quantitative superiority of the proposed method. 
In the synthesized RGB composites, our approach exhibits exceptional structural preservation and chromatic consistency relative to the ground truth.
Whereas LMS and ADWM manifest noticeable blurring along edge discontinuities and hue misalignment in vegetated regions. 
The residual maps reveal that our method generates the darkest error distributions, indicating minimal radiometric deviation.
While competitive methods such as OTPNet and SSDiff produce comparatively brighter residuals, particularly in high-frequency urban textures. 
ARConv demonstrates moderate performance but suffers from over-sharpening artifacts visible as highlighted glows around building perimeters.
Notably, the residual patterns of our approach exhibit homogeneous noise characteristics without systematic bias, suggesting balanced spatial-spectral optimization. 

This visual evidence corroborates the statistical improvements in RR and FR metrics, confirming that the integrated attention-fusion mechanism effectively mitigates spectral distortion while enhancing fine-scale detail reconstruction.
The perceptual alignment with the reference image, combined with the lowest error variance, validates that our model robustly generalizes to complex scenes.
More visual comparison results are in the supplementary material.

\subsection{Ablation Study}

\noindent\textbf{Structure Ablation.}
We conducted a series of comprehensive ablation experiments to thoroughly evaluate the effectiveness of our design. The results from R1 and R2 demonstrate that the introduction of adaptive convolution significantly enhances the performance of the diffusion model (DM). However, this improvement comes at the cost of a substantial increase in computational complexity, highlighting the trade-off between performance and efficiency.

The findings from R2 and R5 provide further evidence of the effectiveness of our single-step distillation strategy. This approach successfully reduces the model's computational burden from 6495 GFLOPs to 130GFLOPs, resulting in an impressive 50x acceleration. Remarkably, this speed-up is achieved while maintaining performance that is nearly identical to that of the original 50-step model across all evaluation metrics (R1). This underscores the effectiveness of our distillation method in optimizing computational efficiency without compromising on model performance.

Additionally, the results from R3, R5, and R8 reveal that integrating the end-to-end model with the single-step diffusion model using lightweight convolutional layers yields a substantial improvement in performance. This integrated approach significantly outperforms both the original end-to-end model and the diffusion model individually. Notably, the computational overhead introduced by these convolutional layers is minimal, with an increase of only 12G FLOPs, which is negligible when compared to the overall computational cost of the model. This suggests that the performance gains are achieved with a relatively small impact on computational resources.

On the other hand, the analysis from R3, R6, R5, and R7 indicates that the addition of extra convolutional layers beyond those used in the integration does not lead to any significant improvement in performance. This suggests that the model reaches a performance plateau after incorporating the optimal convolutional architecture, and further layers do not contribute meaningfully to enhancing the results.

Considering the complexity of Fose, we reduce the FLOPs from 2205 GFLOPs to 297 GFLOPs, achieving a speedup ratio of 7.42.
Moreover, the performance on both RR and FP metrics is improved obviously.

In conclusion, our four-step training strategy successfully enables a lightweight fusion of the diffusion model and the end-to-end model. This fusion significantly boosts the performance of the pansharpening task, demonstrating both efficiency and effectiveness in enhancing the overall model’s capabilities while reducing computational costs.

\noindent\textbf{Hyperparameter Ablation.}
To analyze the impact of architectural depth on OTPNet, we vary the number of blocks in the E2E module and report the results in Tab.~\ref{tab:ablation-depth}.
Among all configurations, the 4-block model achieves the best overall results, with the lowest SAM (2.846) and ERGAS (2.102) and the highest SCC (0.985), indicating the most balanced spatial–spectral reconstruction. 
While using 3 blocks leads to insufficient feature extraction, further deepening the network to 5 or 6 blocks does not yield consistent improvements. 
Instead, it increases computational cost. 
Moreover, when evaluated on full-resolution data without a ground-truth reference, the 4-block model still yields the highest reconstruction quality, confirming its robustness across both reduced-resolution and full-resolution settings. 
These findings demonstrate that a 4-block E2E design provides the most effective and efficient configuration.

\noindent\textbf{Noise Ablation.}
To further investigate the influence of noise initialization, we compare two settings: Zero noise and Random Gaussian noise. 
As shown in Tab.~\ref{tab:ablation-noise}, the Zero noise configuration consistently yields the best reconstruction quality across almost all metrics. 
Specifically, Zero noise achieves the lower SAM (2.783 vs. 3.094) and ERGAS (2.055 vs. 2.130), along with the higher Q8 (0.917) and HQNR (0.933), indicating superior spatial–spectral fidelity. 
In contrast, initializing the process with random noise leads  degraded performance, especially in Q8 and HQNR, suggesting that unnecessary stochastic perturbations harm. 
These results demonstrate that deterministic zero-noise initialization is more suitable, as it provides a clean starting point that stabilizes the distilled model and enhances reconstruction accuracy.
\vspace{-2mm}
\section{Conclusion}
\vspace{-2mm}
In this paper, we propose Fose, a lightweight pansharpening framework, which integrates a one-step distilled diffusion model with an end-to-end network through a compact four-stage training strategy.
By progressively compressing the diffusion process and introducing an efficient fusion mechanism, the proposed design effectively balances spatial detail recovery and spectral fidelity while significantly reducing the computational burden of traditional multi-step diffusion models. 
Experiments demonstrate that our method achieves superior performance compared to existing state-of-the-art approaches in both accuracy and efficiency. 
Overall, this work provides a practical and effective solution for remote sensing pansharpening and for incorporating diffusion priors into lightweight image fusion networks.

{
    \small
    \bibliographystyle{ieeenat_fullname}
    \bibliography{main}
}

\end{document}

%% file: preamble.tex
\usepackage{amsmath}
\usepackage{multirow}
\usepackage{colortbl}
\usepackage{color}
\usepackage{arydshln}
\usepackage{pifont}
\usepackage[ruled,linesnumbered]{algorithm2e}

\newcommand{\cmark}{\ding{51}} % ✓
\newcommand{\xmark}{\ding{55}} % ✗
\definecolor{table_gray}{gray}{0.95}
\definecolor{table_green}{HTML}{acf7c5}